\documentclass{article}

\usepackage{microtype}
\usepackage{graphicx}
\usepackage{subfigure}
\usepackage{booktabs} 

\usepackage{hyperref}
\usepackage{multirow}

\usepackage[accepted]{icml2025}

\usepackage{amsmath}
\usepackage{amssymb}
\usepackage{mathtools}
\usepackage{amsthm}

\usepackage[capitalize,noabbrev]{cleveref}

\theoremstyle{plain}

\theoremstyle{definition}

\theoremstyle{remark}

\usepackage[disable,textsize=tiny]{todonotes}

\icmltitlerunning{Content-Rich AIGC Video Quality Assessment via Intricate Text Alignment and Motion-Aware Consistency}

\begin{document}

\twocolumn[
\icmltitle{Content-Rich AIGC Video Quality Assessment via Intricate Text Alignment and Motion-Aware Consistency}



\icmlsetsymbol{equal}{*}

\begin{icmlauthorlist}
\icmlauthor{Shangkun Sun}{equal,yyy,comp}
\icmlauthor{Xiaoyu Liang}{equal,yyy}
\icmlauthor{Bowen Qu}{yyy}
\icmlauthor{Wei Gao}{yyy,comp}
\end{icmlauthorlist}

\icmlaffiliation{yyy}{SECE, Peking University}
\icmlaffiliation{comp}{PengCheng Laboratory}


\icmlcorrespondingauthor{Wei Gao}{gaowei262@pku.edu.cn}


\vskip 0.3in
]



\printAffiliationsAndNotice{\icmlEqualContribution} 

\begin{abstract}
The advent of next-generation video generation models like \textit{Sora} poses challenges for AI-generated content (AIGC) video quality assessment (VQA). 
These models substantially mitigate flickering artifacts prevalent in prior models, enable longer and complex text prompts and generate longer videos with intricate, diverse motion patterns. 
Conventional VQA methods designed for simple text and basic motion patterns struggle to evaluate these content-rich videos.
To this end, we propose \textbf{CRAVE} (\underline{C}ontent-\underline{R}ich \underline{A}IGC \underline{V}ideo \underline{E}valuator), specifically for the evaluation of Sora-era AIGC videos. 
CRAVE proposes the multi-granularity text-temporal fusion that aligns long-form complex textual semantics with video dynamics. 
Additionally, CRAVE leverages the hybrid motion-fidelity modeling to assess temporal artifacts.
Furthermore, given the straightforward prompts and content in current AIGC VQA datasets, we introduce \textbf{CRAVE-DB}, a benchmark featuring content-rich videos from next-generation models paired with elaborate prompts.
Extensive experiments have shown that the proposed CRAVE achieves excellent results on multiple AIGC VQA benchmarks, demonstrating a high degree of alignment with human perception. 
All data and code will be publicly available. 
\end{abstract}

\begin{figure*}[ht]
    \centering
    \includegraphics[width=1.9\columnwidth]{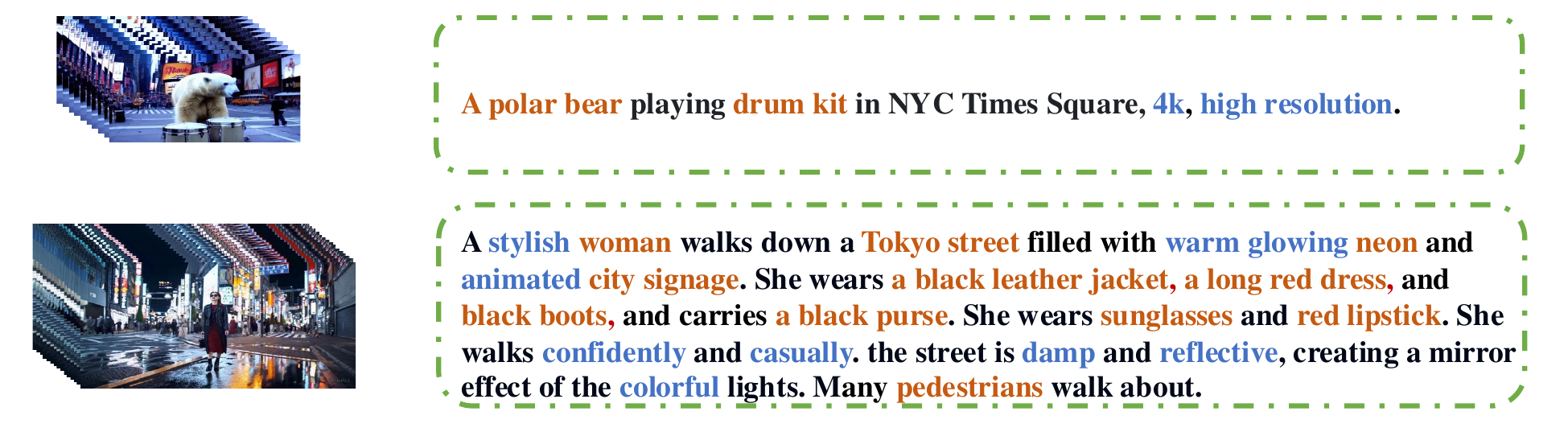} 
    \caption{Comparison of concurrent and previous AIGC videos. Videos are generated by Lavie~\cite{wang2023lavie} (1st row) and Sora~\cite{sora} (2nd row), respectively. Nouns that should be present in the video are highlighted in orange, while adjectives with more details are highlighted in blue. The new-generation AIGC videos contain richer content.}
    \label{fig:1}
\end{figure*}

\section{Introduction}

Recently, text-driven video generation~\cite{sora,hunyuan} has seen significant growth. However, evaluating these text-driven AI-generated videos presents unique and escalating challenges. These challenges primarily stem from two key issues: (1) the need for precise video-text alignment, especially with complex and lengthy text prompts; (2) the occurrence of distinct distortions that are not typically found in naturally generated videos, such as irregular motion patterns and objects.

With the advancement of new-generation video models, these challenges have become even more pronounced. These new-generation models, marked by the advent of Sora~\cite{sora}, offer substantial improvement in generation quality compared to previous models, characterized by rich details and content, such as Kling~\cite{kling}, Gen-3-alpha~\cite{gen3}, Vidu~\cite{vidu}, etc. Compared with previous AIGC videos, these models support \textbf{much longer and more intricate text prompt (often over 200 characters), as well as more complex motion patterns with longer duration (often over 5 seconds with the fps of 24)}. As illustrated in~\cref{fig:1}, these rich contents impose greater demands on the evaluator's ability to understand video dynamics and its relationship with complex textual semantics.

To address this, we introduce Content-Rich AIGC Video Evaluator (CRAVE), to assess the quality of these next-generation text-driven videos. 
CRAVE evaluates videos from three perspectives: It firstly considers the traditional visual harmony like the previous Video Quality Assessment (VQA) method~\cite{dover}, which measures the aesthetics and distortions.
Furthermore, CRAVE leverages a multi-granularity text-temporal fusion module to align the intricate texts with video dynamics. 
Additionally, CRAVE incorporates the hybrid motion-fidelity modeling that exploits hierarchical motion information to assess the temporal quality of the next-generation AIGC videos.

Besides, the gap between the naturalness and complexity of the latest AIGC videos and previous videos has become markedly apparent. 
To better assess current AIGC videos, we introduce CRAVE-DB, a content-rich AIGC VQA benchmark consisting of 1,228 elaborate text-driven videos generated by advanced models such as Kling~\cite{kling}, Qingying~\cite{ying}, Vidu~\cite{vidu}, and Sora~\cite{sora}. Here, by 
"elaborate text," we mean prompts that include complete descriptions of the subject, actions, and environment, with at least \textbf{5} detailed descriptions for any one aspect and a total character count exceeding \textbf{200}. These videos have largely eliminated issues prevalent in previous generations, such as flickering, weak motion, and short content. 
They encompass diverse scenes, subjects, actions, and rich details, with a duration of over 5 seconds and a frame rate of 24 fps. 
Extensive experiments show that CRAVE has achieved leading human-aligned video quality assessment results across multiple metrics on T2V-DB~\cite{t2vqa}, currently the largest AIGC VQA dataset, and the proposed CRAVE-DB.

To summarize, our main contributions are as follows: (1) We introduce CRAVE, the effective evaluator for content-rich videos derived from the new-generation video models, which assess AIGC videos from the temporal and video-text consistency via effective motion-aware video dynamics understanding and multi-granularity text-temporal fusion module. (2) Given the gap between new-generation AIGC videos and previous ones, we introduce CRAVE-DB, a benchmark containing AIGC VQA samples produced by advanced models like Kling, etc., to facilitate the evaluation of contemporary content-rich AIGC videos. (3) Extensive experiments demonstrate the proposed CRAVE achieves excellent results on multiple AIGC VQA benchmarks with varying sources of videos and prompt lengths, showcasing a strong understanding of the quality of AIGC videos.


\section{Related Work}

\subsection{Measurement for Text-to-Video Models}
Currently, common methods used in evaluating text-driven generated videos include some objective metrics~\cite{clip,fvd,is} and human-aligned methods~\cite{pickscore,trivqa,t2vqa}.
Objective metrics such as CLIP-score~\cite{clip} measure the mean cosine similarity between the text and each frame. IS~\cite{is} utilizes the inception feature to measure the overall quality of image and video frames. Flow score~\cite{vbench} calculates dynamic degree via optical flow models such as~\cite{raft,skflow}.
However, these objective metrics do not align with human subjective perception and often evaluate videos from a single dimension. Some methods for evaluating natural video quality provide human-aligned overall evaluations~\cite{dover,fastvqa,stablevqa}. 
DOVER~\cite{dover} assesses quality in terms of aesthetics and technicality. 
FastVQA~\cite{fastvqa} utilizes grid mini-patch sampling to assess videos efficiently while maintaining accuracy.
Q-Align~\cite{wu2023qalign} transforms the video quality assessment task into the generation of discrete quality level words via the Multimodal Large Language Model.
StableVQA~\cite{stablevideo} measures video stability by separately obtaining the raw optical flow, semantic, and blur features.
These methods are suitable for natural video quality assessment but do not consider the text-video alignment, which is key to the evaluation of text-driven videos.
To address this, EvalCrafter~\cite{evalcrafter} assesses quality through a series of indicators including CLIP score, SD score, and natural video quality assessment methods. 
T2V-QA~\cite{t2vqa} incorporates a transformer-based encoder and Large Language Model to assess text-driven AIGC videos.
TriVQA~\cite{trivqa} explores the video-text consistency through cross-attention pooling and the recaption of Video-LLaVA. 
However, there are still relatively fewer VQA methods specifically for AIGC videos. With the growth of new-generation videos, the requirements for understanding video dynamics and text consistency are becoming increasingly demanding, posing greater challenges.

\subsection{Text-to-Video Generation Method}
With the surge of diffusion models~\cite{stablediffusion,ddpm}, a lot of video generation models emerge~\cite{make-a-video,lavie,modelscope,svd,videocrafter, opensora, opensora-plan}.
They represent a significant breakthrough in video generation. However, videos produced by previous methods still tend to suffer from issues such as low resolution, short duration, flickering, and distortion.
With the advent of Sora~\cite{sora}, the new-generation models~\cite{hunyuan, luma, hailuo,tongyi,pika,cogvideox} have made notable progress. Particularly recently, methods like Kling~\cite{kling}, Gen-3-alpha~\cite{gen3}, and Qingying~\cite{ying} have achieved impressive video generation results and have been made available for community testing.
These videos generally alleviate the foundational problems seen in previous methods, with a duration of more than 5 seconds and frame rates above 24 fps.
Meanwhile, the content in these videos includes a lot of details, and they support the control via longer text inputs.
Under the wave of new-generation video generation models, effectively assessing more complex spatiotemporal relationships within the videos and exploring their consistency with longer texts is a topic worthy of further study.

\subsection{Text-to-Video VQA Dataset}
To evaluate and further promote the development of T2V models, some text-to-video VQA datasets have been proposed. 
Despite this, there are still relatively few text-to-video QA datasets suitable for evaluating current AIGC videos.
EvalCrafter~\cite{evalcrafter} collects 700 prompts and uses 5 models to generate 2500 videos in total.
FETV~\cite{liu2023fetv} utilizes 619 prompts to generate 2,476 videos by 4 T2V models. 
Chivileva~\cite{chivileva2023measuring} derives 1,005 videos generated from 5 T2V models.
VBench ~\cite{huang2023vbench} uses nearly 1,700 prompts and 4 T2V models to generate 6984 videos. 
T2VQA-DB~\cite{kou2024subjectivealigneddatasetmetrictexttovideo} contains 10,000 videos generated by 1000 prompts. 
These datasets mainly meet two challenges:
(1) According to the  ITU-standard~\cite{itu}, the number of human annotators should exceed 15 to keep the assessment error within a controllable range.
Among these, only T2VQA-DB~\cite{kou2024subjectivealigneddatasetmetrictexttovideo} and Chivileva~\cite{chivileva2023measuring} meet the standard with 27 and 24 annotators. 
(2) The gap between previous and the concurrent AIGC videos. Previous videos often involve only easy movements and commonly have basic issues such as flickering, which are relatively rarely seen in the new-generation video models.
In this work, to address the issue that previous VQA datasets do not cover the annotated next-generation AIGC videos, we introduce CRAVE-DB, which includes 1,228 next-generation AIGC videos with subjective scores from 29 annotators, to provide a robust assessment of concurrent AIGC videos.

\section{Content-Rich AIGC VQA Benchmark}
With the rapid advancements in text-driven video generation models, the current state-of-the-art models show significant differences compared to previous models in terms of visual quality, content complexity, and the understanding of the input text, as shown in~\cref{fig:1}. These models have substantially alleviated basic issues such as flickering that were prevalent in earlier models, and have removed the length limitation of 77 tokens in CLIP for text-based input found in many past models. 
The challenges now shift towards evaluating content distortion in more complex spatiotemporal scenarios and aligning semantic consistency with more intricate texts.
However, current AIGC VQA datasets still consist of those based on the previous generation of general models, creating a significant gap compared to the concurrent content-rich models. To this end, we introduce CRAVE-DB, a new AIGC VQA benchmark featuring intricate text prompts, content-rich videos generated by state-of-the-art video generation models, and the corresponding human scores. The dataset contains 1,228 videos produced from state-of-the-art video models, incorporating 410 intricate prompts. Each video has a duration of over 5 seconds, with a fps of 24. As for the subjective feedback, each video was scored by 29 human annotators. We will introduce the process of prompt collection, video generation, and subjective study in the following paragraphs.

\begin{figure}[htbp]
\centering
\includegraphics[width=0.6\columnwidth]{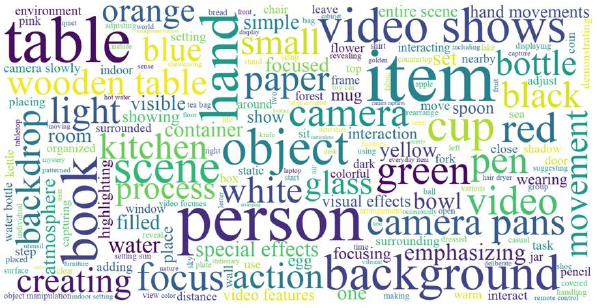}
\caption{Word cloud of prompts in CRAVE-DB.}
\label{fig:word_cloud}
\end{figure}

\begin{table}[]
\setlength\tabcolsep{3pt}
\centering
\caption{Comparison of prompt density and annotator counts.}
\begin{small}
\scalebox{0.95}{
\begin{tabular}{lccc}
\toprule
Dataset & \# Words & \# Chars. & \# Ann. \\ \midrule
FETV~\cite{liu2023fetv}	& 10.94	& 59.60	& 3 \\
EvalCrafter~\cite{evalcrafter}	&12.33	& 69.77	& - \\
T2V-CompBench~\cite{t2vcompbench}	& 10.42	& 56.42	 & 3 \\
VBench~\cite{vbench}	& 7.64	& 41.95	& - \\
VideoGenEval~\cite{vgeneval}	& 33.00	& 202.02	& 0 \\
T2VQA-DB~\cite{t2vqa}	        & 12.32	& 76.22	& 27 \\
\textbf{CRAVE-DB (Ours)}	& \textbf{68.38} & \textbf{411.30}	& \textbf{29} \\
\bottomrule
\end{tabular}
}
\end{small}
\label{tab:prompt_len}
\end{table}

\subsection{Prompt Collection}
\begin{figure}[htbp]
\centering
\includegraphics[width=1.03\columnwidth]{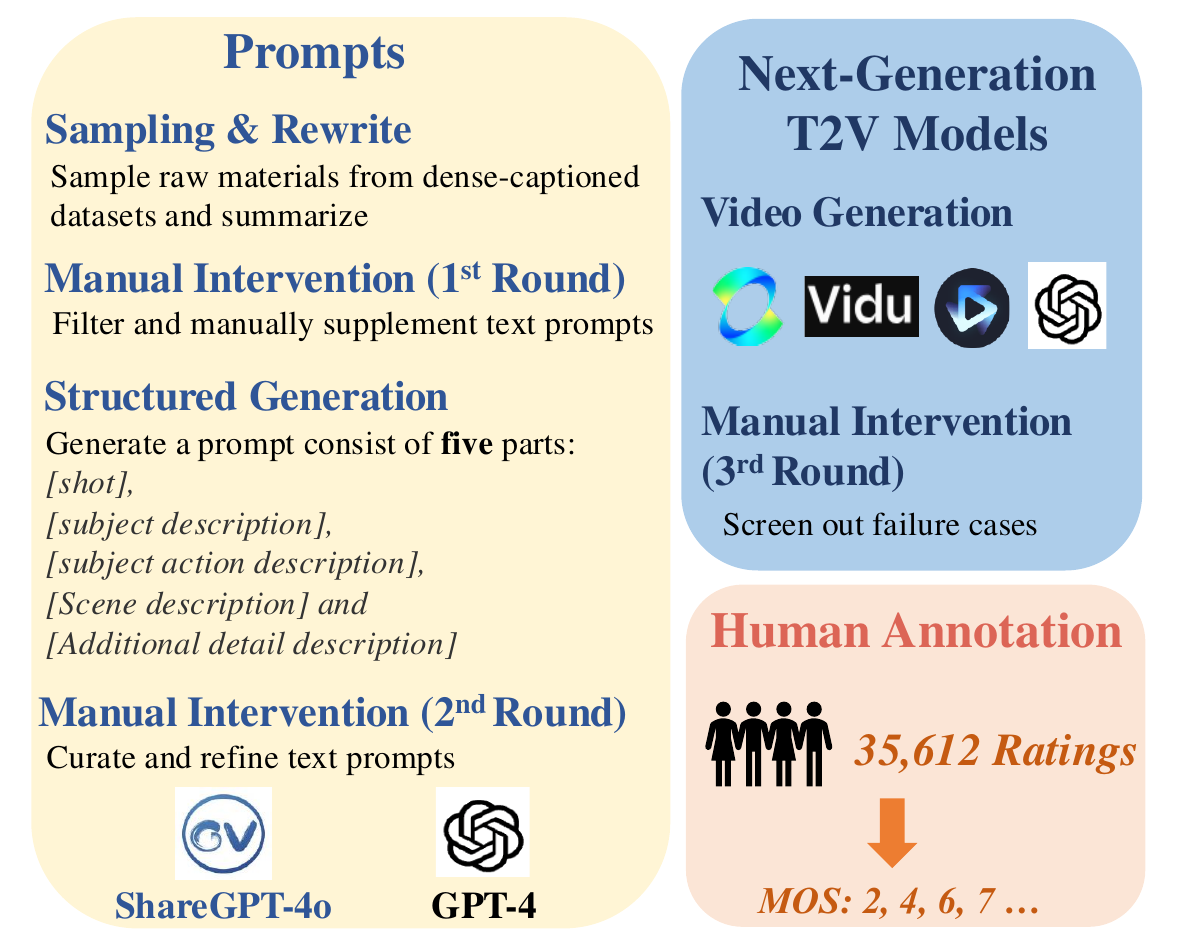}
\vskip -0.1in
\caption{The collection of the proposed CRAVE-DB.}
\vskip -0.1in
\label{fig:cravedb_collection}
\end{figure}

The past AIGC VQA datasets were composed of previous-generation models, where most supported prompt length is limited by CLIP~\cite{clip}. In this case, these prompts tend to be brief, making it challenging to incorporate complex motion descriptions and scene compositions. For instance, we present the prompt density (average word and character count per prompt) of different datasets, as shown in~\cref{tab:prompt_len}. We could learn that most prompts in previous datasets contain merely a dozen words. This inherent limitation poses significant challenges for models when evaluating more sophisticated semantic alignment.

To address this, we propose to construct prompts with richer information. Our overall pipeline is shown in~\cref{fig:cravedb_collection}. To ensure the prompts are detailed and semantically rich, we focus on the prior dense-captioned dataset, ShareGPT-4o dataset~\cite{sharegpt}, which leverages the advanced multimodal capabilities of GPT-4o to describe videos in detail. This dataset contains rich annotations that even require summarization to be clear prompts. We randomly sampled 300 captions from this dataset and summarized them using GPT-4~\cite{gpt4}, retaining only the key details. We then conducted the 1st round of manual intervention to filter out failed, redundant, or illogical generations.

Given that ShareGPT-4o primarily focuses on daily life scenarios, we manually crafted 200 more prompts to broaden the coverage of actions, subjects, and scenes. Prompts contain 4 categories: landscape, object, animal, and human. The ``landscape`` contains common scenes (e.g., grasslands, streets), rare environments (e.g., volcanoes, auroras), and renowned landmarks. The ``animal" includes various mammals, reptiles, birds, fish, and amphibians. The ``object" covers common real-world items, while the ``human" features people across ages, genders, occupations, and clothing.

Subsequently, we employed GPT-4 to structure raw prompts using a template format: ``[shot language] + [subject description] + [subject action description] + [scene description] + [additional detail description]". The ``shot language" incorporates various cinematographic techniques including tilt shots, flat shots, progressive shots, surround shots, close-ups, and panoramic views. The scene descriptions encompass natural landscapes under diverse weather and lighting conditions. Following this, we initiated a second round of manual intervention to screen and refine all prompts, ultimately finalizing a curated set of 410 high-quality prompts. The overall word cloud is shown in~\cref{fig:word_cloud}.

\begin{figure}[htbp]
\centering
\includegraphics[width=1.0\columnwidth]{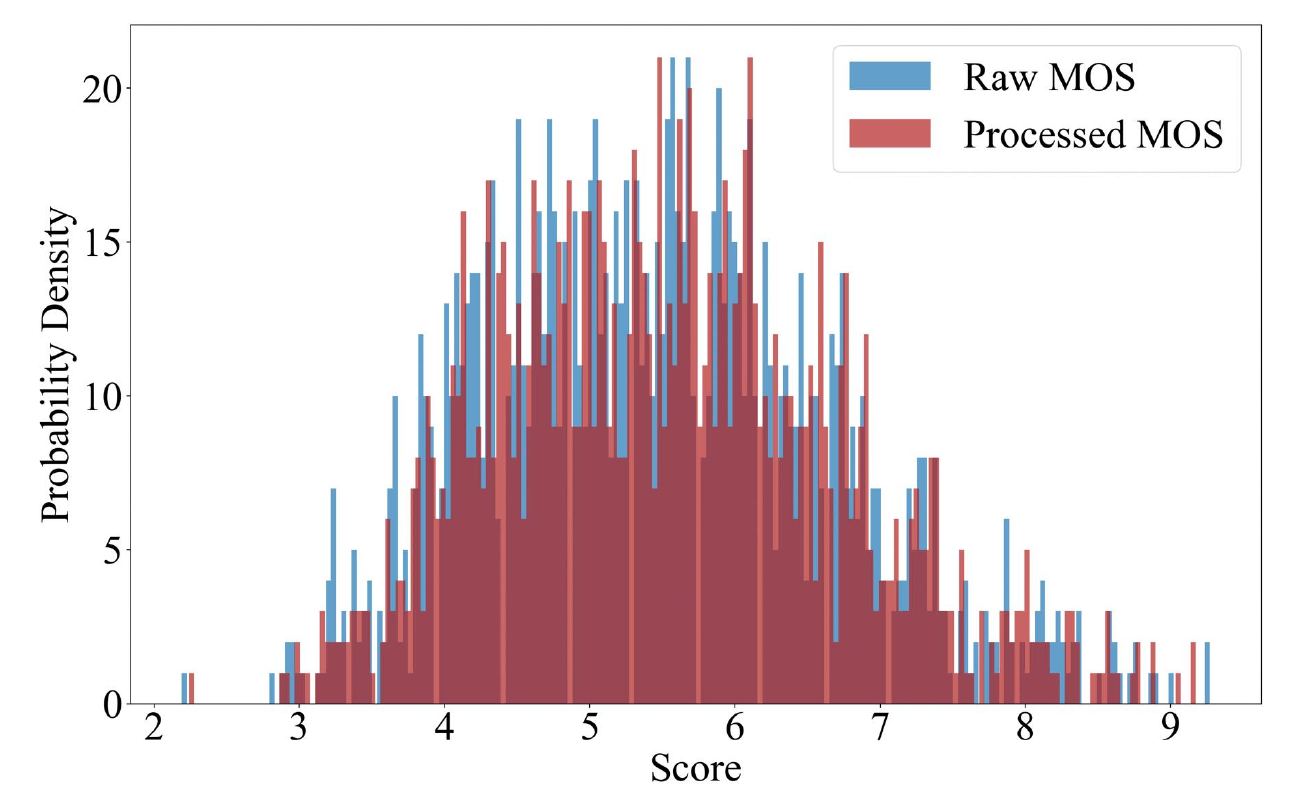}
\vskip -0.1in
\caption{Distribution of MOS in CRAVE-DB.}
\vskip -0.1in
\label{fig:crave_mos}
\end{figure}

\subsection{Video Generation}
\label{sec:video_generation}
Since the advent of Sora~\cite{sora}, text-driven video generation methods have achieved significant advancements in visual quality, text understanding, and the diversity and complexity of generated content. Given the substantial gap between current AIGC videos and prior ones, constructing datasets using the next-generation video models is essential. In this work, we employ Sora and other subsequent state-of-the-art models: Kling~\cite{kling}, Vidu~\cite{vidu}, Qingying~\cite{ying} to build samples. As Sora had not been publicly released by the time of our subjective evaluation, we curated 14 content-rich prompts and their corresponding outputs from Sora’s publicly showcased videos, resulting in a total of 1,228 videos. All videos exceed 5 seconds in duration with a frame rate of 24 fps, and resolutions ranging from $384 \times 688$ to $960 \times 1440$, depending on the generation model. To evaluate newer video generation models and validate CRAVE’s 0-shot generalization capability, we utilized prompts from VideoGenEval~\cite{vgeneval} to generate videos using recently accessible models such as Hunyuan~\cite{hunyuan}, Sora (post-API release), Seaweed Pro~\cite{seaweed}, Mochi 1~\cite{mochi} etc. These outputs were then evaluated using CRAVE, as shown in~\cref{fig:rank}. 
Compared with other datasets, VideoGenEval exhibits relatively higher prompt density (\cref{tab:prompt_len}) and employs distinct prompts compared to CRAVE-DB. Despite its lack of annotations, it is suitable for 0-shot testing.

\begin{figure*}[htbp]
\centering
\includegraphics[width=1.7\columnwidth]{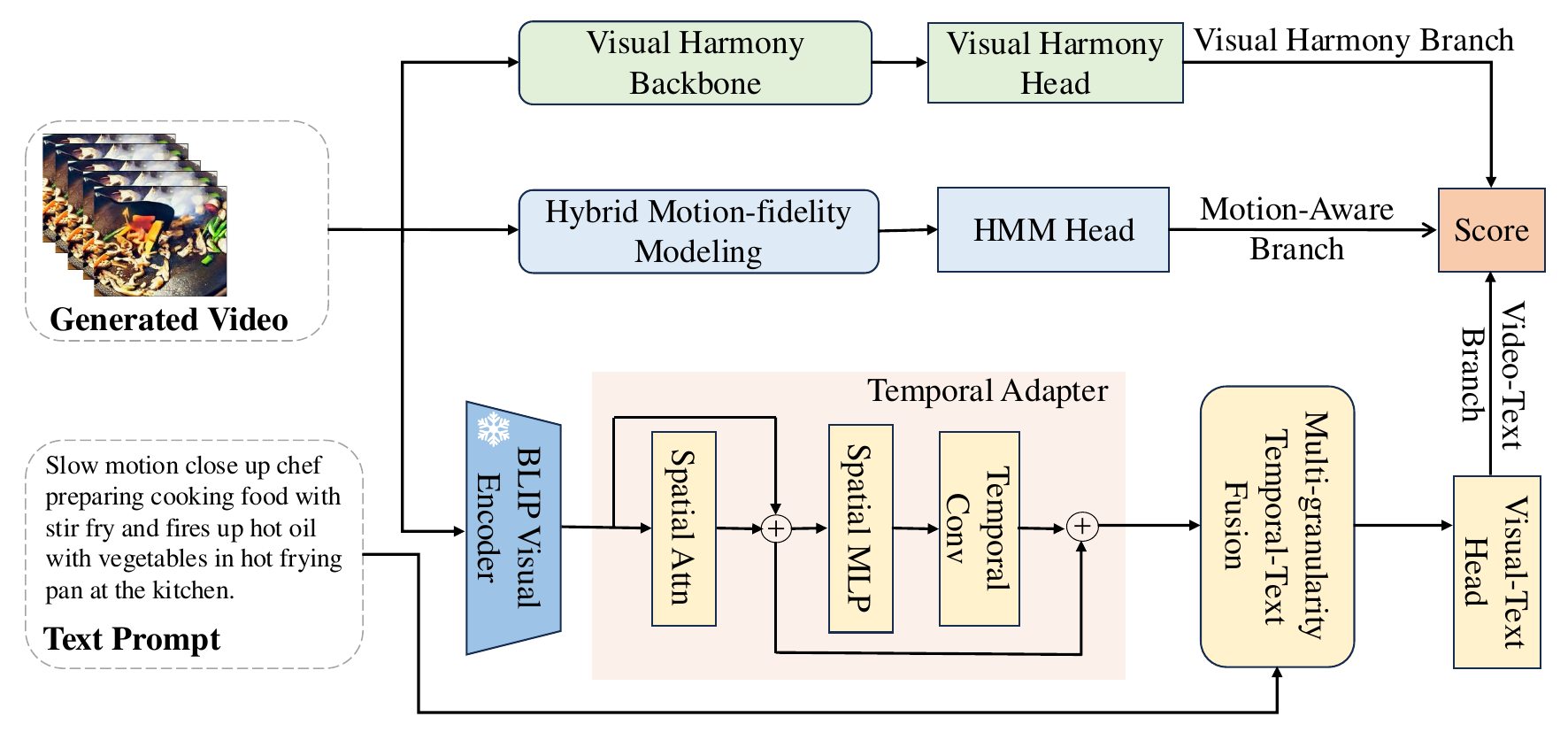} 
\caption{Network overview of the proposed CRAVE.}
\label{fig:crave_pipeline}
\end{figure*}

\subsection{Subjective Study}

According to ITU~\cite{itu} standards, subjective experiments should involve at least 15 participants to reduce error fluctuations.
To obtain the Mean Opinion Score (MOS) for each video, in our experiment, each video was scored by 29 different human subjects. These participants come from diverse backgrounds, including science, engineering, business, law, etc., with all being over 18 years old. Prior to scoring, all participants were gathered on-site for training. During the training, we presented some cases outside the dataset, including good, bad, and average examples, to ensure a basic understanding of the task.
The scoring was conducted on-site, with a mandatory 5-minute break after every 15 minutes of scoring to prevent fatigue. Based on prior work~\cite{vebench,iebench}, the scoring system used a 10-point scale.
During the scoring, the annotators were advised to assess the video from three perspectives: (1) visual quality, which is commonly used in traditional VQA methods; (2) matching between the text and video; and (3) motion quality, such as whether motion consistency is maintained, whether the motion is distorted, and whether it aligns with common sense. Following prior works~\cite{t2vqa, ntire_vqa}, to better estimate the total score of the video, the subjects were asked to provide a final overall score after considering all three aspects. After all the scoring was completed, we obtained a total of 35,612 subjective scores, from which we derived the raw MOS score. As shown in~\cref{fig:crave_mos},  we then normalize the raw scores as Z-score MOS, which could be formulated as:
\begin{align}
    MOS_{z}(m, i) = \frac{X_{m,i} - \mu(X_i)}{\sigma(X_i)},
\end{align}
where $X_{m, i}$ and $MOS_{z}(m,i)$ denotes the raw and Z-score MOS of $m$-th video of $i$-th annotator, respectively. $\mu(\cdot)$ and $\sigma{(\cdot)}$ refer to the mean and standard deviation function, respectively. $X_{i}$ represents all MOS from the $i$-th annotator. After deriving $MOS_{z}(m, i)$, the screening method in BT.500~\cite{bt500} is deployed to filter the outliers.

\section{Content-Rich AIGC Video Evaluator}
\subsection{Overall Framework}
CRAVE evaluates content-rich AIGC videos from three perspectives:
(1) visual harmony, measured using traditional video quality metrics like aesthetics and distortion, (2) text-video semantic alignment, achieved via Multi-granularity Text-Temporal (MTT) fusion, and (3) motion-aware consistency, which is specific for dynamic distortions in AIGC videos, captured through Hybrid Motion-fidelity Modeling (HMM).
The overall framework is illustrated in~\cref{fig:crave_pipeline}. We will delve into the details of each module in the following sections.

\begin{figure}[t]
\centering
\includegraphics[width=1.05\columnwidth]{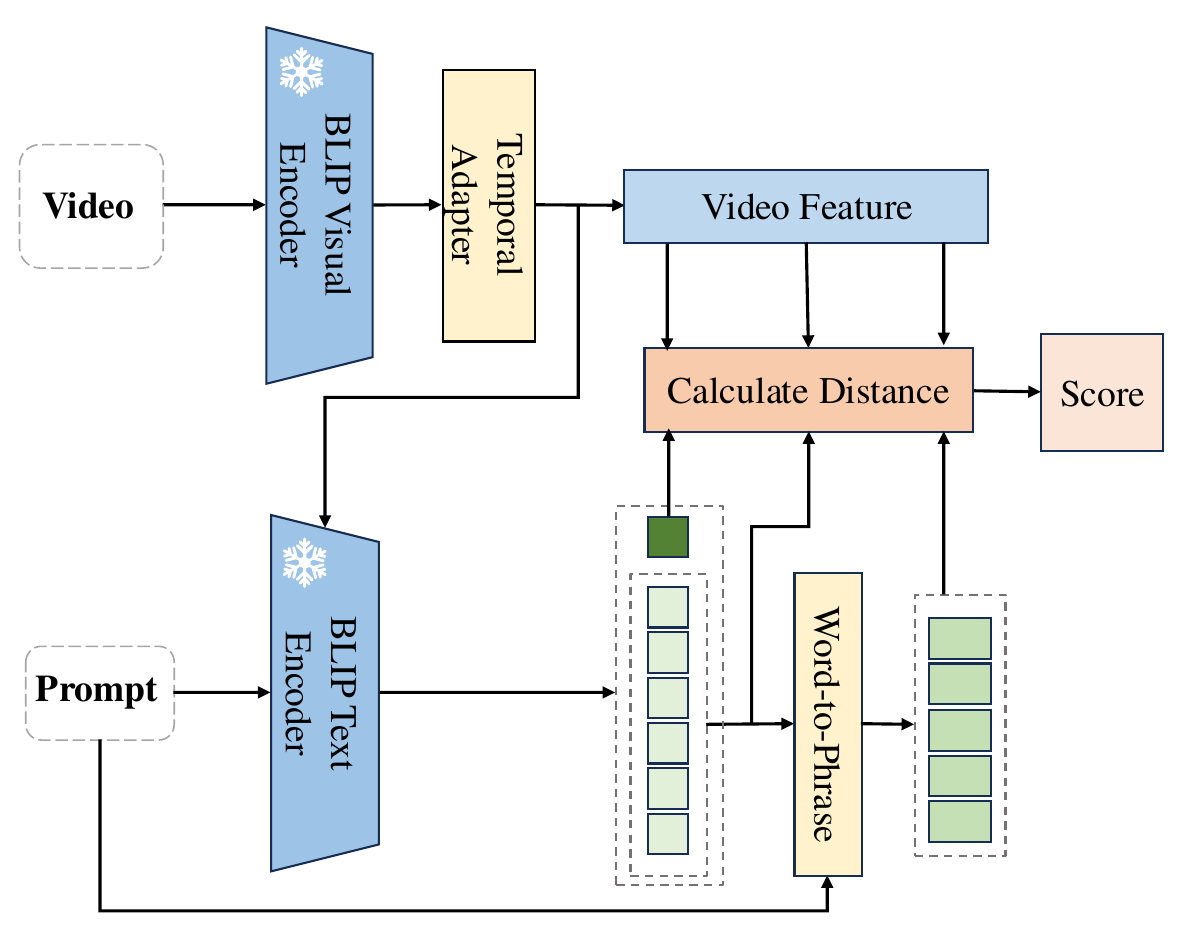} 
\caption{Details of the proposed MTT module for text alignment.}
\label{fig:mtt}
\end{figure}

\subsection{Visual Harmony}
For traditional natural video quality assessment, we utilize DOVER~\cite{dover} to assess individual videos from the aesthetic and technical perspectives given its success. DOVER evaluates videos based on two dimensions: aesthetic score and technical distortion. In this work, we use the pre-trained DOVER method as the Visual Harmony Backbone, followed by a linear head to obtain the output, which could be formulated as:
\begin{align}
    F_{aes} &= \Phi_{aes}(V), \\
    F_{tech} &= \Phi_{tech}(V), \\
    O_{vh} &= \omega_{vh}(F_{aes} \oplus F_{tech})),
\end{align}
where $V$ is the input video, $\Phi_{aes}$, $\Phi_{tech}$ represent the aesthetic encoder and distortion encoder in DOVER, respectively, $\oplus$ denotes concatenation along the dimension, $\omega_{vh}$ represents the linear head for this branch, and $O_{vh}$ refers to the corresponding output.

\subsection{Multi-Granularity Text-Temporal Fusion}
Multi-granularity Text-Temporal (MTT) module firstly leverages high-quality priors from multi-modal understanding approaches like BLIP~\cite{li2022blip}, successfully extending it in the temporal dimension via effective temporal adapters.
Then the visual information aggregated in the temporal adapter interacts with the text embedding via cross-attention. 
To flexibly fuse effective information from the text, we additionally perform multi-granularity aggregation on the text. 
Namely, in addition to the whole input, we break down the text into phrases and words of varying granularity containing different levels of semantic information via SpaCy~\cite{spacy}. After that, the integrated text embeddings of varying granularity are measured with the output of the visual branch, as shown in~\cref{fig:mtt}. 
The entire process can be formulated as: 
\begin{align}
    F_v &= \Phi_{t}(\Phi_{s}(V)), \\
    F_e &= \Phi_{e}(Concat([enc],F_v), e), \\
    F_{\text{enc}} &= F_e[0, \ldots], \\
    F_{\text{word}} &= F_e[1:, \ldots],
\end{align}
where $V$, $e$, $F_v$ are the input video, text prompt, and the derived visual feature, respectively. 
$\Phi_{s}, \Phi_{t}, \Phi_{e}$ denote the spatial encoder, temporal adapter, and text encoder, respectively.
$F_{\text{enc}}$ and $F_{\text{word}}$ refer to the $[enc]$ token embedding and the word token embedding, representing different levels of semantics of the textual prompt.
To get the phrase-level encoding for prompts, we utilize the successful word-to-phrase module in~\cite{zhu2023deep}.
The module converts the mapping between word and phrase to the mapping between word embedding and phrase embedding, based on the position mapping between word and the corresponding embedding, which could be formulated as,
\begin{align}
    \{p_1,p_2\ldots,p_m\} &= \Phi_{w}(\{w_1,w_2\ldots,w_n\}), \\
    \{F_{p_1},F_{p_2}\ldots,F_{p_m}\} &= \Phi_{F_w}(\{F_{w_1},F_{w_2}\ldots,F_{w_n}\}),
\end{align}
where $\Phi_{w}$ and $\Phi_{F_w}$ refer to the mapping between word and phrase and the mapping between word embedding and phrase embedding. $p_i$, $w_i$ refers to the $i$-th useful phrase and word in prompt, while $F_{p_i}$, $F_{w_i}$ refers to the $i$-th  phrase embedding and word embedding, respectively.

After that, we calculate the cosine distance between all text features and video features respectively. Finally, we sum them up to get the final score.
\begin{align}
    O_{align} &=\sum_{l}{cos(F_l,F_v)},
\end{align}
where $l$ denotes different levels of granularity such as the whole paragraph level, phrase level and word level.

\subsection{Hybrid Motion-fidelity Modeling}
Compared with natural videos, AIGC videos usually contain unique distortions such as irregular objects and motions that violate the physical laws. Despite significant improvements in recent video generation models, low-fidelity motion remains a persistent challenge. Here, motions that defy logic, deformed motions, and motions with abnormal amplitudes are collectively referred to as "low-quality" motions. To better assess motion distortions in current AIGC videos, we propose Hybrid Motion-fidelity Modeling (HMM), which hierarchically captures motion features at different granularities. Specifically, considering the successful application of optical flow in anomaly detection~\cite{caldelli2021optical,agarwal2020detecting}, we leverage dense motion information derived from optical flow to capture low-level motion patterns, combined with global abstract motion information from action recognition tasks~\cite{kinetics, somethingsomething}. The experimental section later demonstrates the effectiveness of combining these two aspects. In practice, flow features are extracted using the pre-trained StreamFlow~\cite{streamflow}, while high-level abstract motion priors are obtained from the pre-trained Uniformer~\cite{uniformer}. Different branches are then input into the feed-forward network and pass through a linear head to regress and obtain the final output. 

\subsection{Supervision}
Based on the previous training objective in ~\cite{dover,fastvqa,simplevqa}, the mix of rank loss~\cite{rankloss} and PLCC (Pearson Linear Correlation Coefficient) loss are adopted to train the overall network. Note that the BLIP visual and text encoder remains frozen throughout the training process. The whole training objective could be formulated as,
\begin{align}
    \mathcal{L} &= \mathcal{L}_{plcc} + \gamma \cdot \mathcal{L}_{rank},
\end{align}
where $\gamma$ is the coefficient set to 0.3 in the experiments.

\section{Experiments}
\subsection{Implement Details}
We use T2VQA-DB~\cite{t2vqa} and the proposed CRAVE-DB for evaluation. 
T2VQA-DB is the largest AIGC VQA dataset for text-driven video generation. It contains plenty of AIGC videos from previous methods, which provides a good complement to the evaluation. 
When training on T2VQA-DB, we use the 10-fold cross-validation, which randomly partitions the whole dataset into 10 equal-sized folds and uses 9 folds for training and 1 fold for testing. We follow the training settings in DOVER~\cite{dover} and models are trained for 20 epochs with linear probing and then fine-tuned with full parameters for 10 epochs. For CRAVE-DB, 
we train 40 epochs on the training split and evaluate on the test set. We use the Adam optimizer with the initial learning rate of $1e-3$ and batch size of 8.

\begin{table}[h]
\setlength\tabcolsep{3pt}
\centering
\caption{Quantitative comparison on CRAVE-DB.}
\begin{small}
\scalebox{0.9}{
\begin{tabular}{llccc}
\toprule
Type & Models  & SRCC & PLCC & KRCC \\ \midrule
\multirow{12}{*}{0-shot}
& UMTScore~\cite{liu2023fetv} & 0.0134	& 0.0355 & 0.0118  \\
& Flicker~\cite{vbench} & 0.0761 & 0.0687 & 0.0493 \\
& Bg. Consis.~\cite{vbench} & 0.1170 & 0.0829 & 0.0773 \\
& Sub. Consis.~\cite{vbench} & 0.1418 & 0.0956 & 0.0936 \\
& ViCLIP~\cite{viclip} & 0.1290 & 0.1207 & 0.0857 \\
& LongCLIP~\cite{longclip}  & 0.2757 & 0.3033 & 0.1884 \\
& Aes. Score~\cite{vbench} & 0.3557 & 0.3457 & 0.2373 \\
& HPSv2~\cite{hpsv2} & 0.0562 & 0.0501 & 0.0391 \\
& PickScore~\cite{pickscore} & 0.0557 & 0.0492 & 0.0361 \\
& ImageReward~\cite{xu2023imagereward} & 0.2216 & 0.2125 & 0.1470 \\
& Motion Sm.~\cite{vbench} & 0.2331 & 0.2630 & 0.1549 \\
\midrule
\multirow{6}{*}{Ft.} 
& SimpleVQA~\cite{simplevqa} & 0.6230 & 0.6180 & 0.4436 \\
& StableVQA~\cite{stablevqa} & 0.6396 & 0.6415 & 0.4552 \\
& FastVQA~\cite{fastvqa} & 0.7155 & 0.7062 & 0.5243 \\
& DOVER~\cite{dover} & 0.7095 & 0.7192 & 0.5224 \\ 
& TriVQA~\cite{trivqa} & 0.7183	& 0.7313 & 0.5293 \\ 
& T2VQA~\cite{t2vqa} & 0.7266  & 0.7098 & 0.5369  \\
\midrule
Ours 
& CRAVE & \textbf{0.7587} & \textbf{0.7581} & \textbf{0.5660} \\
\bottomrule
\end{tabular}
}
\end{small}
\label{tab:2}
\end{table}

\subsection{Quantitative Results}
As shown in~\cref{tab:1} and~\cref{tab:2}, we can learn that CRAVE achieves leading performance on both the proposed content-rich dataset and the T2VQA-DB, which includes prior AIGC videos. On CRAVE-DB, CRAVE demonstrates a particularly significant lead, highlighting its effectiveness in evaluating next-generation AIGC videos. 
On T2VQA-DB, CRAVE also outperforms previous models, even surpassing LLM-based models such as Q-Align and T2VQA, which further demonstrates the effectiveness of its multidimensional design. ``Ft." denotes methods that need further fine-tuning on the target dataset. ``Bg.", ``Sub.", ``Consis", ``Aes.", ``Sm." denote the background, subject, consistency, aesthetic, and smoothness, respectively. 
0-shot methods tend to have lower results, which is also observed in previous works~\cite{t2vqa, vebench}. It could be due to the lack of alignment with human perception or the consideration of the dynamic distortion in AIGC videos.

\begin{table}[]
\setlength\tabcolsep{3pt}
\centering
\caption{Quantitative comparison on T2VQA-DB.}
\scalebox{0.88}{
\begin{tabular}{llccc}
\toprule
Type & Models  & SRCC & PLCC & KRCC \\ \midrule
\multirow{5}{*}{0-shot}
& CLIPSim~\cite{clip} & 0.1047 & 0.1277 & 0.0702 \\
& BLIP~\cite{li2022blip} & 0.1659 & 0.1860 & 0.1112 \\
& ImageReward~\cite{xu2023imagereward} & 0.1875 & 0.2121 & 0.1266 \\
& ViCLIP~\cite{viclip} & 0.1162 & 0.1449 & 0.0781  \\ 
& UMTScore~\cite{liu2023fetv} & 0.0676 & 0.0721 & 0.0453 \\ \midrule
\multirow{5}{*}{Ft.} 
& SimpleVQA~\cite{simplevqa} & 0.6275 & 0.6338 & 0.4466 \\
& BVQA~\cite{bvqa} & 0.7390 & 0.7486 & 0.5487 \\
& FastVQA~\cite{fastvqa} & 0.7173 & 0.7295 & 0.5303 \\
& DOVER~\cite{dover} & 0.7609 & 0.7693 & 0.5704 \\
& Q-Align~\cite{wu2023qalign} & 0.7601 & 0.7768 & 0.5860 \\
& T2VQA ~\cite{t2vqa} & 0.7965 & 0.8066 & 0.6058  \\ \midrule
Ours 
& CRAVE & \textbf{0.8122} & \textbf{0.8214} & \textbf{0.6338} \\
\bottomrule
\end{tabular}
}
\label{tab:1}
\end{table}

\subsection{Qualitative Results}
We visualize the difference between the predicted and the ground-truth MOS, as shown in the supplements. The curves are obtained from a fourth-order polynomial nonlinear fitting. We further present the scores of different AIGC videos assessed via CRAVE in the supplements. 

\subsection{Zero-Shot Ranking Comparison}
\begin{figure*}[htbp]
\centering
\includegraphics[width=1.99\columnwidth]{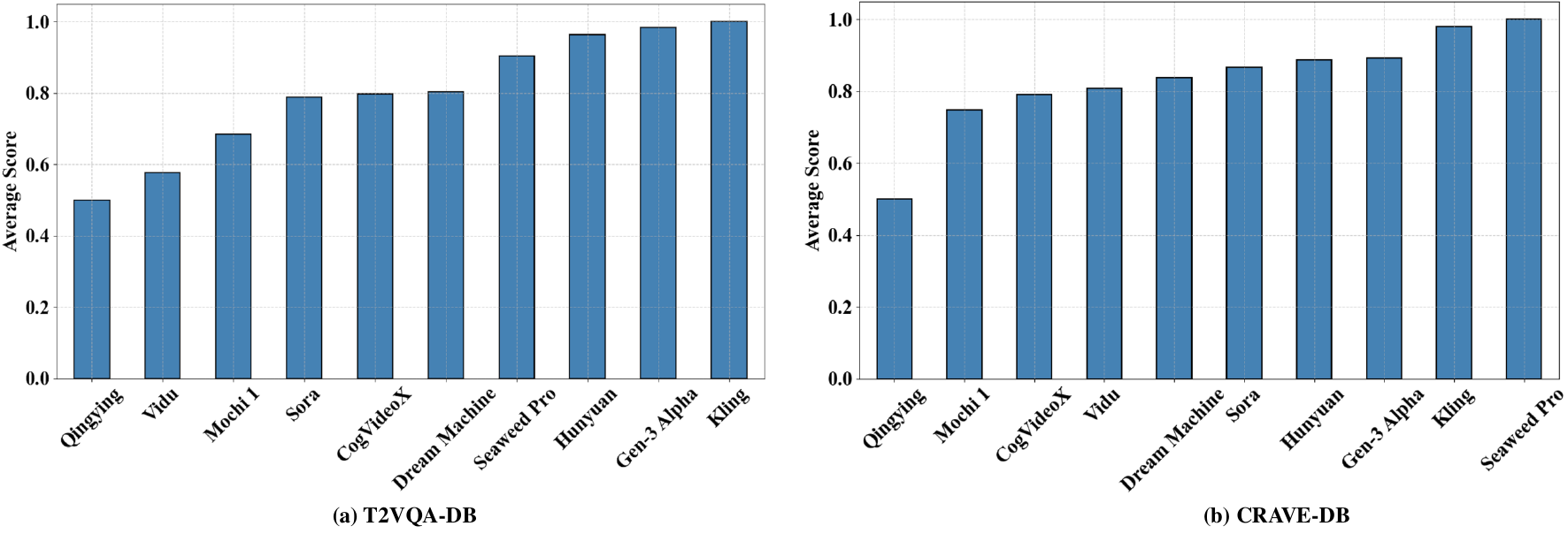} 
\caption{The ranking of next-generation models provided by models trained on different AIGC VQA datasets.}
\label{fig:rank}
\end{figure*}

We present the rankings of next-generation video generation model scores evaluated by CRAVE after training on different VQA datasets. As discussed in~\cref{sec:video_generation}, VideoGenEval~\cite{vgeneval} was selected for this experiment due to its relatively high prompt density, its completely distinct data sources compared to CRAVE-DB, and its inclusion of newer models. We utilized all 424 text-to-video (t2v) prompts and generated results from VideoGenEval, which include recent models such as~\cite{ying,vidu,mochi,sora,cogvideox,luma,seaweed,hunyuan,gen3,kling}. As shown in ~\cref{fig:rank}, (a) and (b) correspond to CRAVE using the pretrained weights in~\cref{tab:1} and~\cref{tab:2}, respectively.

\begin{table*}[h]
\centering
\caption{Quantitative results of ablation study.}
\begin{small}
\begin{tabular}{llcccccc}
\toprule
\multirow{2}{*}{Experiment} & \multirow{2}{*}{Method} & \multicolumn{3}{c}{CRAVE-DB} & \multicolumn{3}{c}{T2VQA-DB} \\ \cmidrule(lr){3-5} \cmidrule(lr){6-8}
 & & SRCC & PLCC & KRCC & SRCC & PLCC & KRCC \\ \midrule
\multirow{4}{*}{Temporal-Text Fusion} &
None & 0.6512	& 0.6601	& 0.4643 & 0.7701 & 0.7804 & 0.5885 \\
& ST-Graph. & 0.6966	& 0.6968	& 0.5042 & 0.7990 & 0.8084 & 0.6196 \\ 
&Temp. Attn. & 0.7159	& 0.7175	& 0.5226 & 0.7989 & 0.8089 & 0.6199  \\
& \underline{Pseudo 3D Conv} & \textbf{0.7310} & \textbf{0.7241} & \textbf{0.5335} & \textbf{0.8077} & \textbf{0.8167} & \textbf{0.6289}  \\ \midrule
\multirow{5}{*}{Multi-Granularity Text Injection} &
None & 0.7477	& 0.7458	& 0.5526 & 0.8077 & 0.8167 &0.6289 \\
&Global & 0.7520	& 0.7485	& 0.5566 & 0.8115 & 0.8207 & 0.6329 \\ 
&Local & 0.7524	& 0.7498	& 0.5578 & 0.8121 & 0.8212 & 0.6337 \\ 
&Individual & 0.7491	& 0.7460	& 0.5534 & 0.8111 & 0.8200 & 0.6326 \\
&\underline{Combined} & \textbf{0.7587}	& \textbf{0.7581}	& \textbf{0.5660} & \textbf{0.8122} & \textbf{0.8214} & \textbf{0.6338} \\ \midrule
\multirow{4}{*}{Motion-Aware Modeling} &
None & 0.7507	& 0.7499	& 0.5561 & 0.8103 &0.8199 &0.6317 \\ 
& High-Level & 0.7527	& 0.7529	& 0.5605 & 0.8111 & 0.8204 &0.6326 \\ 
& Low-Level & 0.7529	& 0.7513	& 0.5576 & 0.8118 & 0.8211 &0.6330 \\ 
& \underline{Hybrid} & \textbf{0.7587}	& \textbf{0.7581}	& \textbf{0.5660} & \textbf{0.8122} & \textbf{0.8214} & \textbf{0.6338} \\ \midrule 
\multirow{3}{*}{Flow Frames}
& 4 & 0.7562	& 0.7552	& 0.5598 & 0.8110 & 0.8206 & 0.6323 \\ 
& 8 & 0.7566	& 0.7564	& 0.5612 & 0.8113 & 0.8208 & 0.6326 \\ 
& \underline{16} & \textbf{0.7587}	& \textbf{0.7581}	& \textbf{0.5660} & \textbf{0.8122} & \textbf{0.8214} & \textbf{0.6338} \\ 
\bottomrule
\end{tabular}
\end{small}
\label{tab:3}
\end{table*}

\subsection{Ablation Study}
To verify the effectiveness of the proposed method, we ablate each component on the design of CRAVE, as shown in~\cref{tab:3}. Underlined settings are used in our final model.
We conducted experiments on both CRAVE-DB and T2VQA-DB. As CRAVE-DB naturally contains intricate texts, rich motion information, and other such content, We could learn that the improvements on CRAVE-DB are generally more significant. We first explore ways to align the text with temporal visual information. ST-Graph, namely the spatiotemporal graph, flattens the time dimension into the spatial dimension for calculation. Temp. Attn. denotes the attention along the additional temporal dimension. Pseudo 3D Conv is inspired by~\cite{make-a-video}, where additional convolutions are stacked in the temporal dimension. We can see that the effectiveness has significantly improved with temporal modeling, and that the Pseudo 3D Conv widely used in generation tasks also excels in long-text spatiotemporal modeling. We further investigate the granularity of MTT and discover that integrating all granularity levels yields optimal performance. Additionally, we examine the impact of motion-aware temporal modeling. 
Our experiments demonstrate that dense data from optical flow enhances overall performance, and incorporating sparse abstract spatiotemporal information provides a significant performance boost. We further explored the impact of flow frames on the results. Specifically, we calculated the optical flow using 4 frames, 8 frames, and 16 frames during the process. We observed that using more optical flow frames tends to improve accuracy. Given the trade-off between accuracy and efficiency, we ultimately chose 16 frames for the flow calculation.

\section{Conclusion}
Given the gap between current AIGC videos and AIGC VQA dataset, we introduce CRAVE, an effective VQA method, and CRAVE-DB, a new benchmark for the next-generation AIGC videos. Based on the effective multi-dimensional design, CRAVE achieves excellent human-aligned results across multiple metrics and datasets. CRAVE-DB contains more content-rich prompts and detailed content, along with extensive human annotations, making it closer to the concurrent text-driven AIGC videos.

\section*{Impact Statement}
This paper presents work whose goal is to advance the field of Machine Learning. There are many potential societal consequences of our work, none which we feel must be specifically highlighted here.

\bibliography{example_paper}
\bibliographystyle{icml2025}

\newpage
\appendix
\onecolumn
\section{Qualitative Results.}

\begin{figure*}[htbp]
\centering
\includegraphics[width=1.0\columnwidth]{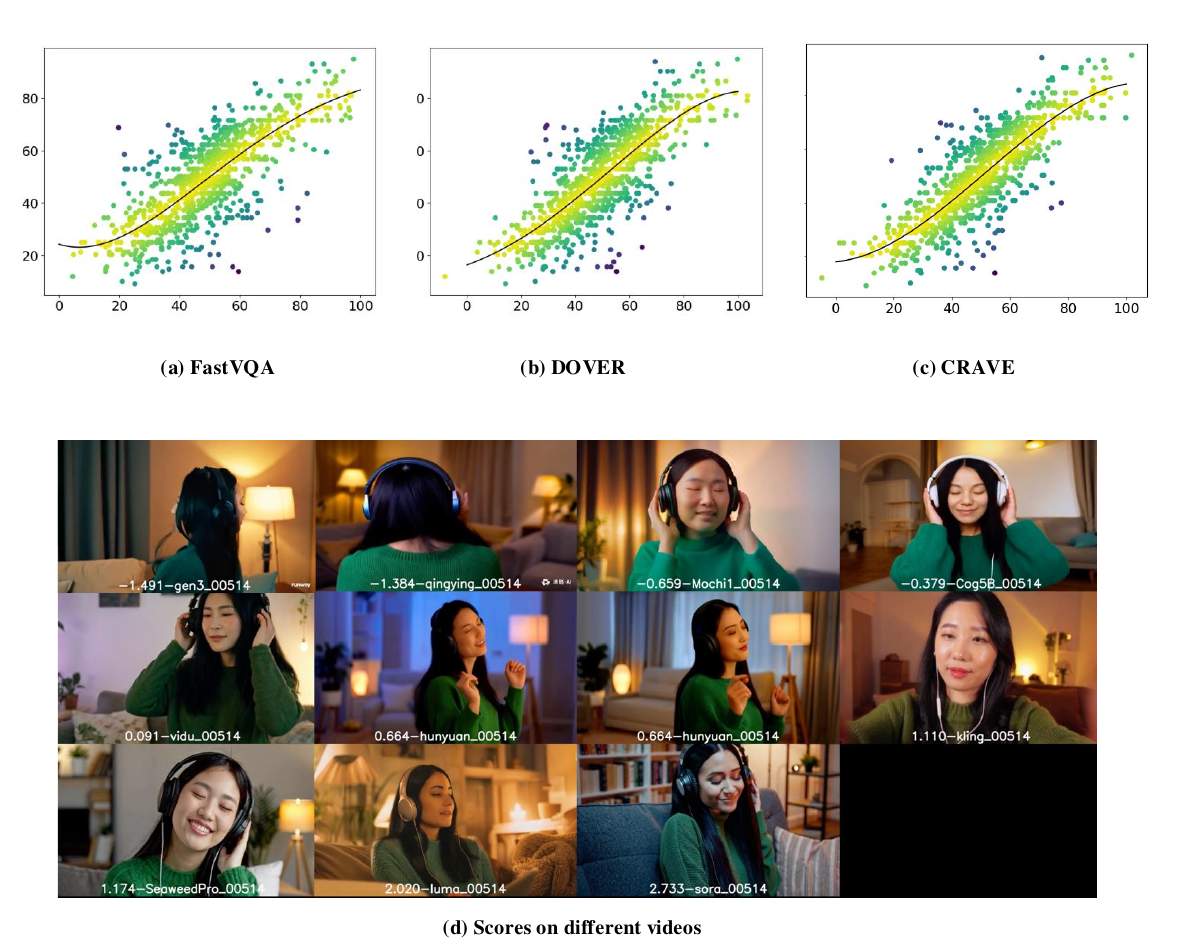} 
\caption{Scatter plots of the predicted scores and ground-truth MOSs. A brighter scatter point represents higher density.}
\label{fig:mos}
\end{figure*}

As shown in ~\cref{fig:mos}, (a), (b), and (c) represent the visualization of the differences between different models on T2VQA-DB. The more clustered the points, the smaller the differences. We can observe that the points in (a) and (b) are more scattered and farther from the central line. A fourth-order polynomial nonlinear fitting is used to draw the central line. (d) shows the scores of CRAVE for the generated results of different models. More detailed videos can be found in the supplementary materials.
Here, the direct output of CRAVE has not been normalized, so negative values may occur.

\end{document}